\documentclass[journal]{IEEEtran}
\usepackage{graphicx}
\usepackage{makecell}
\usepackage{times}
\usepackage{mathrsfs}
\usepackage{epsfig}
\usepackage{graphicx}
\usepackage{amsmath}
\usepackage{amssymb}
\usepackage{graphicx}
\usepackage{cite}
\usepackage{enumerate}
\usepackage{cases}
\usepackage{multirow}
\usepackage{verbatim}
\usepackage{amssymb}
\usepackage{CJK}
\usepackage{algorithm}
\usepackage{algorithmicx}
\usepackage{algpseudocode}
\usepackage{color}
\usepackage{bm}
\usepackage{booktabs}
\usepackage[colorlinks,linkcolor=red]{hyperref}
\usepackage{lineno,hyperref}
\usepackage{tikz}
\usepackage{ifthen}
\usepackage{bm}

\ifCLASSINFOpdf
\else
\fi
\begin{document}
%
\title{Feedback Chain Network For Hippocampus Segmentation}
%
%
%

\author{Heyu Huang, Runmin Cong, Lianhe Yang, Ling Du, Cong Wang, and Sam Kwong
\thanks{Heyu Huang, Lianhe Yang, and Ling Du are with School of Computer Science and Technology, Tiangong University, Tianjin 300387, China (e-mail: huangheyu22@gmail.com).}
\thanks{Runmin Cong is with the Institute of Information Science, Beijing Jiaotong University, Beijing 100044, China (e-mail: rmcong@bjtu.edu.cn).}
\thanks{Cong Wang is with the Distributed and Parallel Software Lab, Huawei Technologies, Shenzhen 518129, China (e-mail: wangcong64@huawei.com).}
\thanks{Sam Kwong is with the Department of Computer Science, City University of Hong Kong, Hong Kong SAR, China, and also with the City University of Hong Kong Shenzhen Research Institute, Shenzhen 51800, China (e-mail: cssamk@cityu.edu.hk).}
}

\maketitle

\begin{abstract}
The hippocampus plays a vital role in the diagnosis and treatment of many neurological disorders. Recent years, deep learning technology has made great progress in the field of medical image segmentation, and the performance of related tasks has been constantly refreshed. In this paper, we focus on the hippocampus segmentation task and propose a novel hierarchical feedback chain network. The feedback chain structure unit learns deeper and wider feature representation of each encoder layer through the hierarchical feature aggregation feedback chains, and achieves feature selection and feedback through the feature handover attention module. Then, we embed a global pyramid attention unit between the feature encoder and the decoder to further modify the encoder features, including the pair-wise pyramid attention module for achieving adjacent attention interaction and the global context modeling module for capturing the long-range knowledge. The proposed approach achieves state-of-the-art performance on three publicly available datasets, compared with existing hippocampus segmentation approaches. The code and results can be found from the link of \url{https://github.com/easymoneysniper183/sematic_seg}.
\end{abstract}


%
\IEEEpeerreviewmaketitle

%
%
%
%

\section{Introduction}
The hippocampus is a brain structure that governs long and short-term memory and spatial orientation, which plays a key role in the diagnosis and treatment of many neurological disorders, such as Alzheimer's disease \cite{prince2015world}, schizophrenia \cite{styner2004boundary,tanveer2020machine}, and major depression \cite{bremner2000hippocampal}. When people are affected by these brain dysfunctions, the clinical manifestations are usually different degrees of atrophy in their shape and size \cite{madan2017emotional,saribudak2018gene}. Early detection of hippocampal volume and morphology changes through 3D imaging with magnetic resonance techniques \cite{suk2015latent,wang2019hand} can better the treat and control of such neurological disorders and prevent their further deterioration. Therefore, accurate and robust segmentation of hippocampal images is essential in clinical practice. However, to date, the manual segmentation method of the hippocampus is difficult to meet the requirements of segmentation accuracy, and it is very subjective and time-consuming. Therefore, researchers hope to realize the automatic segmentation of the hippocampus with the help of image processing related technologies. 

In recent years, with the rapid development of convolutional neural networks (CNNs) \cite{crm/tetci22/PSNet,litjens2017survey,crm/acmmm21/bridgenet,crm/nips20/CoADNet,long2015fully,crm/tmm22/blindSR,zhou2020discriminative,crm/aaai20/GCPANet}, numerous deep models have been successfully leveraged to medical field \cite{zotti2018convolutional,crm/tip20/MCMT-GAN,crm/acmmm20/NuI-Go,lian2018multi,crm/tim22/covid,liu2020superpixel,crm/tce22/covid,crm/jbhi22/polyp}. The methods represented by U-Net \cite{ronneberger2015u} and its variants have become the pragmatic standard in the field of medical image segmentation, which greatly improves the performance of medical image segmentation by bridging the paths among the encoder and decoder stages. In addition,
the multi-scale skip connection \cite{ibtehaz2020multiresunet,lin2021residual}, dilated convolution \cite{liu2018brain,punn2020inception}, and dense block \cite{li2018h,yang2021densely} are embedded into the U-Net framework \cite{Zhu2019DilatedDU} to further improve the segmentation accuracy. However, due to the low contrast between the hippocampus and surrounding tissues, irregular shapes, unobvious edges, and large individual differences, it is very difficult to accurately and completely segment the morphological structure of the hippocampus. Therefore, the hippocampus segmentation network derived from U-Net still needs to solve the following issues: (1) Due to the uniqueness and difference of features at different levels, simply combining semantic and appearance information from coarse/low to fine/high in the inference process lacks the exploration of more and deeper roles of different levels of features on segmentation targets, and it is insufficient; (2) For the hippocampus segmentation task, the intensity of surrounding background is large, which will cause greater interference, and the hippocampus region has a relatively long range, so how to effectively suppress the background and ensure the integrity of the detection results requires further exploration.

To remedy the above problems, we propose a multi-level feedback chain network to achieve hippocampus segmentation, which integrates the feedback chain structure unit and global pyramid attention unit. In the feedback chain structure unit, we design hierarchical feature aggregation feedback chain to capture the deeper and wider feature representation of the corresponding layer. Inspired by Girum et al. \cite{Girum2021LearningWC} emphasizing the important role of feedback manipulation, but in contrast to their looped form, FAFC is composed of two chains with different roles: the main chain and the side chain. The main chain performs deeper extraction and selection of important features, and finally generates semantic expression at this level as feedback. The side chain performs wider aggregation of cross-level complementary features at different levels. In such a chain structure, we can maximize the extraction of hippocampal discriminative features while mitigating the effects of feature dilution. Meanwhile, we propose a feature handover attention module to refine the feedback feature. The high-level semantic features obtained by the feature aggregation feedback chain are fed back to the original backbone feature of the corresponding layer, and then the high-level semantic features and the original backbone feature are effectively combined. In the inference process, feature handover attention module continuously highlights the spatial and semantic responses to guide segmentation. Finally, we embed the global pyramid attention unit between the feature encoding and decoding, which aims to refine the encoder features and capture the global contextual information. On the one hand, the pair-wise pyramid attention module achieves adjacent attention interaction in a multi-scale pyramid structure between adjacent scales. On the other hand, the global context modeling module correlates the features at different locations, thereby eliminating the interference of high similarity categories and constraining the integrity of the segmentation result.

Our main contributions can be summarized as follows:
\begin{itemize}
	\item We propose a hippocampal segmentation network includes the feedback chain structure unit and global pyramid attention unit. Our network achieves state-of-the-art performance against other competitors on three popular benchmarks.
	\item The feedback chain structure unit builds a feature aggregation feedback chain structure including the main and side chains to learn the deeper and wider feature representation of the corresponding layer. Moreover, with the feature handover attention module, by refining and combining feedback information to the backbone feature, the chain structure allows for better feature selection and feedback, and feature aggregation across layers.
	\item The global pyramid attention unit is proposed to restore the encoder features and capture the global contextual information, which includes a pair-wise pyramid attention module that achieves adjacent attention interaction, and a global context modeling module that captures the long-range correlation.
\end{itemize}

\section{Related work}

\subsection{Hippocampus Segmentation}

Before deep neural networks were widely applied to medical image segmentation, most of the best automatic hippocampal segmentation methods used multi-atlas approaches \cite{artaechevarria2009combination,rousseau2011supervised,zarpalas2014accurate,cardenas2019adaptive}. Until today, the method proposed by FreeSurfer \cite{fischl2012freesurfer} is still followed for segmentation of brain structures in medical research. Although its segmentation quality is high, it is too time-consuming. With the rise of deep learning, many time-saving and efficient methods for hippocampal segmentation have emerged. Thyreau et al. \cite{thyreau2018segmentation} proposed the Hippodeep model in 2018, which performed fast planar analysis in the region of interest (ROI) by transferring algorithmic knowledge. Ataloglou et al. \cite{ataloglou2019fast}  proposed segmentation and error correction modules in convolutional neural networks, which used segmentation masks to crop and correct more important segmentation regions to improve the segmentation quality. Folle et al. \cite{folle2019dilated} proposed a new training model in 2019, which was based on 3D U-Net incorporating dilation convolution and deep supervision to improve hippocampal segmentation by obtaining multi-scale information. Cui et al. \cite{cui2018hippocampus} in 2019 based on a 3D densely connected convolutional network to analyze the hippocampus using local visual features and global structural features, resulting in enhanced classification capabilities. Cao et al. \cite{Cao2017MultitaskNN} proposed a multi-task deep learning (MDL) method for joint segmentation and clinical score regression of hippocampal MRI. Liu et al. \cite{Liu2020AMD} proposed a new 3D convolution-based dense model for more comprehensive and detailed acquisition of hippocampal features. Shi et al. \cite{Shi2021DiscriminativeFN} proposed a hierarchical and comprehensive attention approach to extract discriminative features from the hippocampus.

\subsection{Multi-Level Aggregation}

For semantic segmentation, the aim is to accurately classify each pixel in an image. The key factors affecting the results is the processing of detail and semantic information. In other words, to
improve performance, high-resolution feature maps with strong semantic
information need to be obtained. The efficient approaches are to fuse multi-level feature maps, integrating low-resolution strong semantic feature maps with high-resolution spatial feature representations \cite{he2019dynamic,crm/tgrs22/RRNet,he2019adaptive,crm/tcyb22/rsi,kirillov2019panoptic,crm/tip21/DAFNet,zhu2019asymmetric,yuan2019spatial,crm/tgrs19/rsi}. However, how to integrate as well as get effective results is the goal of the competition to explore. PSPNet \cite{zhao2017pyramid} proposed to model multiple scales using the Pyramid Pooling Module (PPM), while the DeepLab series \cite{chen2017deeplab,chen2017rethinking} used astrous spatial pyramid pooling (ASPP) to indirectly obtain multiple scales for feature aggregation. ICNet \cite{zhao2018icnet} set the input as multiscale images and used a cascade approach to aggregate features for efficiency. However, for adjacent
scale aggregation in a single direction, it can continuously increase the information loss of semantic features in propagation, which leads to ineffective propagation of semantics from deep to shallow layers. At the same time, as the depth of the network increases, high-level semantic features are diluted. These are not conducive to the extraction and retention of hippocampal discriminative features. In contrast, our proposed feature aggregation feedback chain and the chain structure it forms can alleviate the above problems. In the main chain, deeper learning and selection of features are performed for each level. In the side chain, the different scales of features are aggregated and connected across levels. Such an approach accomplishes multi-level feature aggregation maximally and resists the effects of feature dilution. In this way, we can obtain discriminative features for the hippocampus.

\begin{figure*}[!t]
	\centering
	\includegraphics[width=\linewidth]{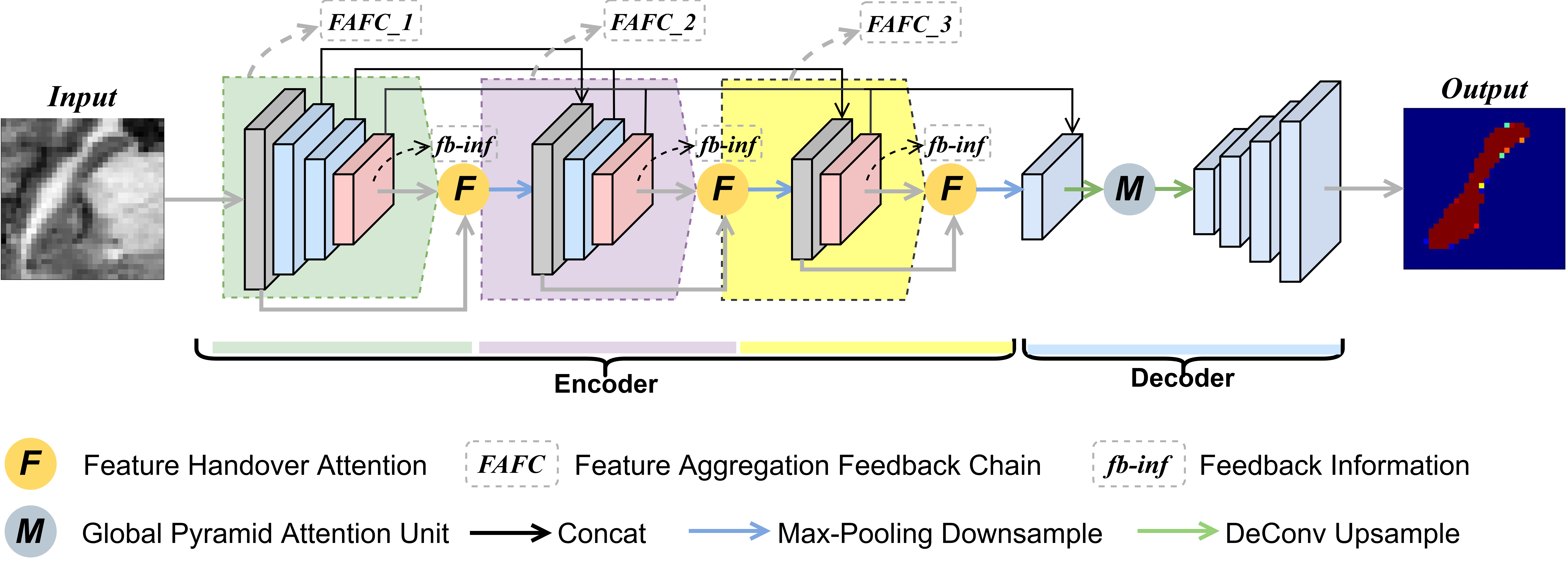}
	\caption{Overall architecture of the proposed feedback chain network. In the figure, the three manifestations of FAFC in the encoding stage are highlighted in green, purple and yellow rectangular boxes. Among them, the gray arrows indicate the main chain processes used for the same level of depth extraction. The black arrow represents the side chain process used for cross-level width aggregation. In addition, the pink feature map represents the feedback feature information obtained from the main chain, which is combined with the encoded backbone features represented by the gray feature map under the role of the FHA module, and jointly carried out in further feature inference.}
	\label{1}
\end{figure*}

\subsection{Attention Mechanism}

Attention mechanisms have received extensive attention in the field of medical image segmentation and other vision tasks \cite{crm/tip22/CIRNet,crm/tip21/DynamicRGBDSOD,crm/tcyb22/glnet,crm/tmm22/TNet,crm/tcsvt22/weaklySOD,crm/tetci22/PSNet,crm/acmmm21/CDINet,crm/CVPR21/depthSR}. It can effectively highlight the useful parts and suppress the redundant information. Many works
have attempted to embed attention modules into network architectures to improve performance. Hu et al. \cite{hu2018squeeze} proposed a squeeze and excitation (SE) module for the attention module, which focused on channel relationships and performed dynamic channel-wise feature realignment to enhance feature representation. In contrast to the use of attention mechanisms to reweight important information, other works have focused on capturing the long-range dependencies of the global environment. Fu et al. \cite{fu2019dual} proposed a dual-attention module consisting of spatial and channel attention to implement semantic segmentation, where the dual-attention module was similar to a non-local operation. Although these works used attention mechanisms to highlight important features and suppressed background interference to some extent. However, the results of attention for these methods were generated independently of the corresponding feature level. For small targets such as the hippocampus, their emphasis on local or overall contour details of the hippocampus was inadequate in response. Therefore, we take into account the relationship between multi-level features and their attention flows in the feature handover attention module. It constantly highlights spatial and strong semantic information between different levels to guide segmentation, which positively affects the final accurate prediction. In addition, we also complete the interactions of adjacent scale attention maps in the global pyramid attention unit, to strongly suppress background noise and high similarity categories. Besides, it captures high responsive global contextual information to improve the integrity of segmentation.

\section{Proposed method}
\subsection{Overview}

In general, we divide the proposed model into three stages, as shown in Fig. \ref{1}, the feedback chain structure (FAFC) for encoding, the global pyramid attention unit (GPA) in the middle and the decoder at the end. The algorithmic structures of the first two stages are closely integrated, firstly, the chain structure is used to achieve parallelized reasoning through feature handover attention module (FHA), replacing the one-way or serial computational process of previous work, which fully resists the effect of feature dilution and retains enough spatial structure information. Also the feedback structure constantly emphasizes high-level important information at different levels, which is an aspect of hippocampal discriminative feature acquisition; secondly, the GPA unit is utilized in an auxiliary role to ensure multiple fine-grained and global responses, which makes the subsequent decoding process more robust. As for the decoding side of dimension recovery, we no longer deliberately pursue its complexity. Such an encoder-heavy but decoder-light framework meets our purpose and is experimentally verified to be effective.

\subsection{Feedback Chain Structure Unit}
\subsubsection{The structure of Feature Aggregation Feedback Chain}
\ 
\newline
\indent The encoding part of the standard U-Net structure extracts the features of the current layer by convolution and transfers them to the next layer to form a multi-level feature representation. In order to further enrich the features of each layer and enhance the feature representation capability of the hippocampus based on dense connection and feature reuse  \cite{woo2018cbam,huang2017densely}, we design the FAFC for each layer, including a main chain and several side chains. The main chain
is for deeper feature learning, and the side chain is for wider feature learning. Specifically, the main chain takes the backbone as the core, and strives to learn deeper important features within the corresponding level. Meanwhile, the main chains receive the information of the pervious main chain via the side chain with dense connection, thereby correlating the main chain features across levels and obtaining wider important features. The output of each layer corresponding to the chain structure contains richer multi-level and cross-level information, which is fed back to the initial backbone features of this layer and integrated with each other with the FHA module, so as to obtain more discriminative features and pass them to the next layer. At this point, a reticular encoder structure equipped with a feedback chain learner is formed. The proposed chain structure has two advantages: (1) Since important features can be diluted as the network deepens, the FAFC can not only enhance the extraction of discriminative features, but also resist the effect of feature dilution. (2) In addition, in this repeated aggregation across levels, the FAFC achieves multiple evaluations of important features for higher-order spatial relations, thus maximizing the retention of discriminative features in the hippocampus and facilitating the accurate acquisition of segmentation results.

Specifically, for a feature $f_n^i$  at a certain position in the encoding process, it aims to use the side chains to achieve parallel computation between different levels (the same scales at different levels correspond to summation) and to use the main chain to complete the information feedback of the semantic features at the tail of each level. The former allows full preservation of details and spatial structure, and the latter allows multiple highlighting of semantic information, which make up the structure of extraction of discriminative features from the hippocampus. In particular, when $i = 1$, $f_n^i$ represents a feature at a certain position on the backbone. In other cases, $f_n^i$ represents a feature at a certain position in the FAFC. The feature map can be formulated as:
\begin{equation}
\footnotesize
f_n^i=\left\{\!{\begin{array}{*{20}{l}}{\hat f_n^{i\!-\!1}+\Omega _n^i(\hat f_n^{i\!-\! 1}),n=1}
	\\{[\hat f_n^{i\!-\!1},\bar f_{n\!-\!1}^i]+\Omega _n^i(\hat f_n^{i\!-\!1},\bar f_{n\!-\!1}^i),n\ge2,i>1 }
	\\{[\hat f_{n\!-\!1}^i,\hat f_{n\!-\!1}^c,\bar f_{n\!-\! 1}^{i\!+\!1}]+\Omega _n^i\!(\bar f_{n\!-\!1}^{i\!+\!1})+\Phi _n^i\!\;(\hat f_n^{i\!-\!1},\hat f_{n\!-\!1}^c), n\ge2,i=1}
	\end{array}}\!\!\!\!\right.
	\end{equation}
where $\hat f_n^i$ denotes the feature map from the main chain, and $\bar f_n^i$ denotes the feature map from the side chain, $n$ means the index of the level and $i$ means the index of the depth in the level, $c$ is the index of the maximum depth in the ${n^{th}}$ level. ${\Omega _n^i}$ is a level process, which includes convolution and concatenation operations, and ${\Phi _n^i}$ is a cross-level process, which includes the processing of the FHA module and the max-pooling operation.
\subsubsection{Feature Handover Attention Module}
\begin{figure}[!t]
	\centering
	\includegraphics[width=\linewidth]{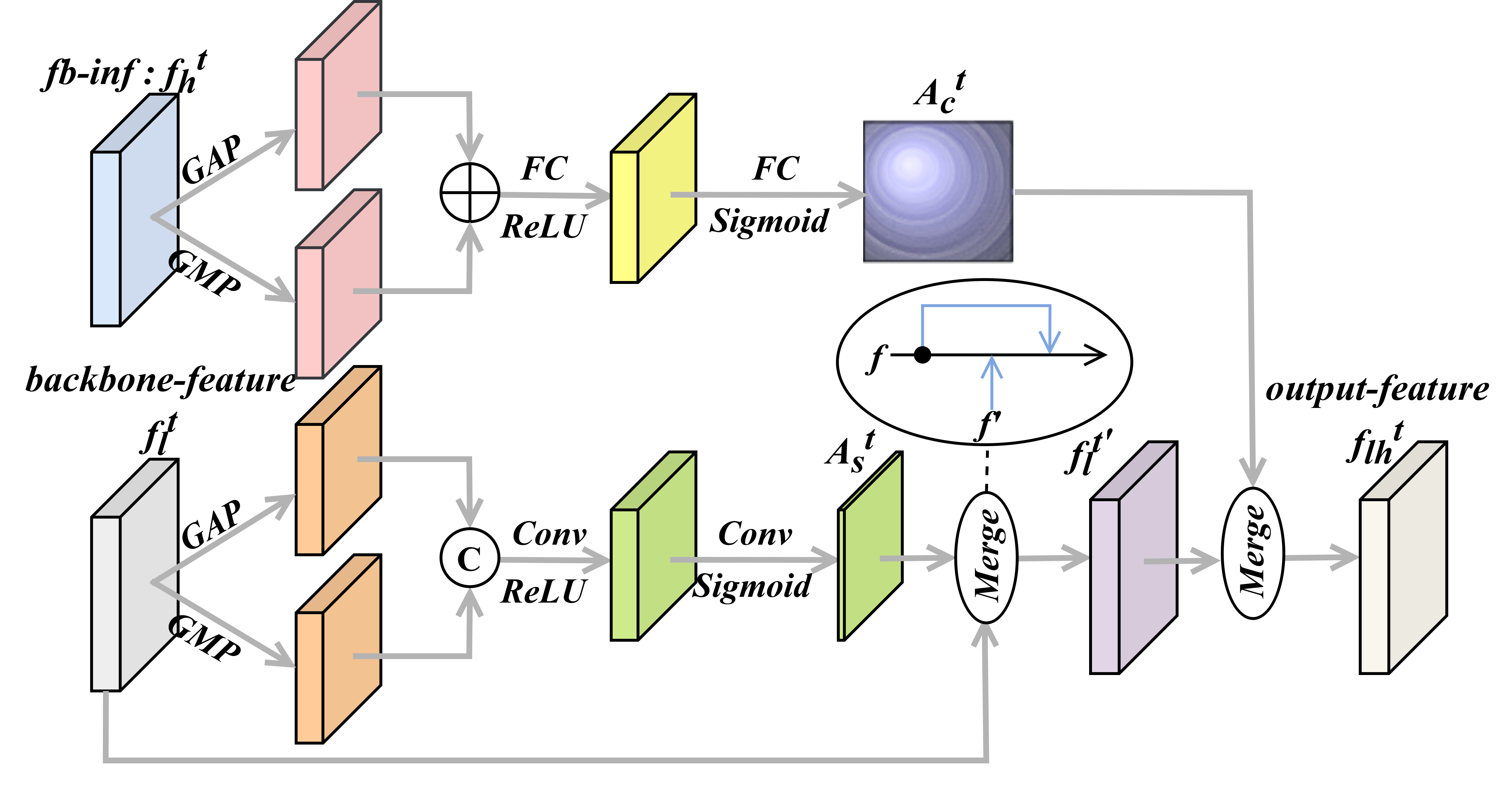}
	\caption{Details of the FHA module. '$C$' indicates the channel concatenation, and '$+$' represents the element-wise addition, '$ Merge $' operation contains both multiplication and addition of its own residuals, '$\mathit{fb}\textit{-}\mathit{inf}$' refers to feedback information.}
	\label{2}
\end{figure}

\ 
\newline
\indent With the feedback features embedded in high-level and deeper semantic information, how to effectively combine with the original backbone features is the main issue that needs to be addressed in the next step. The usual operation is a simple fusion through concatenation or addition, but it lacks further exploration of the relationship between different features. The attention mechanism plays a powerful role in feature selection and highlighting, and proper utilization of it can bring great benefits. Therefore, we design a FHA module to inherit the original backbone features from the current level and the feedback features generated by FAFC. The overall structure of the FHA module is shown in Fig. \ref{2}. Among them, the upper branch refers the initial backbone feature, and the lower branch means the feedback feature. The backbone features have detailed spatial location highlighting, while the feedback features carry semantic channel information prominence.

First, an effective spatial attention method  is applied to the low-level feature $f_l^t$, i.e., the backbone feature of the corresponding level, to obtain an updated weighted feature map $f_l^t{}'$, which focuses more on the foreground region and emphases the spatial structure details, i.e., more structural details of the hippocampus are recovered to make its boundary contours with the background clearer and to suppress the background and boundary noise \cite{woo2018cbam} to some extent. Specifically, we perform average-pooling and max-pooling opreations on the channel axis of $f_l^t$, and concatenate two outputs to generate a spatial attentive descriptor $\Omega _s^t \in {R^{2 \times {H_t} \times {W_t}}}$ which is further by convolution, a 2D spatial attention map $A_s^t \in {R^{{H_t} \times {W_t}}}$ is obtained. Then the attention map $A_s^t$ is multiplied with $f_l^t$ and finally concatenated to $f_l^t$ itself with residual connection to get the updated feature $f_l^t{}'$, which can be formulated as:
\begin{equation}\begin{array}{l}
		A_s^t=Attn(f_l^t)=\sigma (conv\,((\Omega _s^t){\rm{;}}\;\hat \theta))
		\\ \;\;\quad=\sigma (conv(concat(avepool(f_l^t),maxpool(f_l^t)){\rm{;}}\;\hat \theta ))
\end{array}\end{equation}
\begin{equation}
	{f_{l}^{t}}'=f_{l}^{t}\otimes A_{s}^{t}\oplus f_{l}^{t}
\end{equation}
where $t \in \left\{ {1,2,3} \right\}$ indexes the convolution stage, $\otimes$ denotes element-wise multiplication with channel-wise broadcasting, $\oplus$ represents the element-wise summation, $conv( \cdot ;\;\hat \theta )$ are parameters ${\hat \theta }$ corresponding to the convolutional layer, $avepool$ and $maxpool$ are the average-pooling and max-pooling, $concat$ represents feature channel concatenation.

Then, we also apply the effective channel attention method to the high level feature $f_h^t$, i.e., the feedback feature of the corresponding level, in a similar operation, yielding an attention map $A_c^t$, where $A_c^t \in {R^{{C_t}}}$, with high responsive emphasis on the channels of discriminative features of the target, which is precisely the strong response to the rich semantic information from the feedback feature within the level. Considering it as a mask, makes the combination with the above updated feature map $f_l^t{}'$ using multiplication, which corresponds to selection and feedback of important semantic features within the level. It can maximally highlight and supervise the hippocampal discriminative features. Such selection and emphasis in each level is exactly what is needed for the segmentation results, and to guide the segmentation process. Finally, the $f_{lh}^t$ is obtained by forming a residual connection with $f_l^t{}'$ as the final output of the whole module. In this way, the final output $f_{lh}^t$ obtained is then fed into the backbone for the next stage of inference. The above process can be described as:

\begin{equation}\begin{array}{l}
		A_c^t=Attn(f_h^t)=\sigma (conv((\Omega _c^t){\rm{;}}\;\hat \theta))\\
		\quad\;\;=\sigma (conv(concat(avepool(f_h^t),maxpool(f_h^t)){\rm{;}}\;\hat \theta ))
\end{array}\end{equation}
\begin{equation}
	f_{lh}^{t}={f_{l}^{t}}'\otimes A_{c}^{t}\oplus {f_{l}^{t}}'
\end{equation}

In summary, each level of inference ends with FHA module for processing and then converges to the backbone.
\subsection{Global Pyramid Attention Unit}
For the hippocampus segmentation in MRIs, the surrounding background regions often have strong intensity interference, and how to further effectively suppress them needs to be further explored. Hence, we aim to give a global and comprehensive attention response to the feature stream under the feedback chain structure at the appropriate location, between the encoder and decoder. For our proposed GPA unit, it is the additive and auxiliary unit for the high-level important semantics obtained in the encoding part, combining the strong semantics with the spatial contextual highlighting. Therefore, it is also a guarantee for the pixel and spatial dimension recovery in the decoding part.

Specifically, its applications in attention flow are described in two aspects. In the fine-grained aspect, we multiply two-by-two feature layers with different scales of spatial attention masks alternately, so that they obtain different scales of spatial correspondence with each other, and then integrate them over the last scale feature layer in a progressive manner, completing the fine-grained consideration; in the global aspect, we want to preserve as much as possible edge blurred, tiny volume of the complete structure of the hippocampus target, so we dimensionally transform and multiply the above feature matrix to achieve spatially contextualized attention response. Two modules from two different perspectives, i.e., the PPA module and the GCM module, are composed in the GPA unit as a whole.

\begin{figure}[!t]
	\centering
	\includegraphics[width=\linewidth]{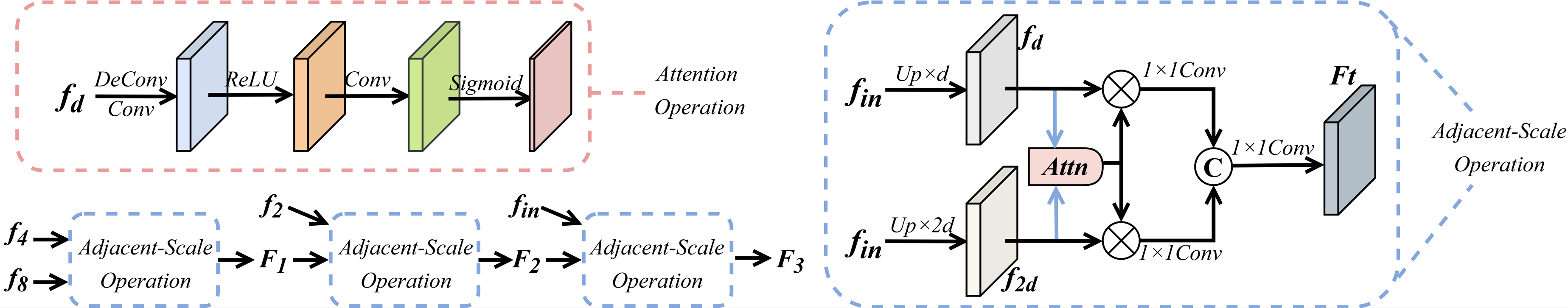}
	\caption{Details of the PPA module. In the figure, the top left corner depicts the process of extracting the attention mask in the '$Attn$' operation, the right side depicts the convergence process of specific neighboring scale features, and the bottom left corner expresses the process of the whole PPA module.}
	\label{3}
\end{figure}
\subsubsection{Pair-wise Pyramid Attention Module}
\ 
\newline
\indent Previous works \cite{zhang2020dense,wang2019salient,yuan2019spatial} demonstrate that the multi-scale pyramid structure combine with attention mechanisms can effectively resist the changes of the object size, and assist in suppressing the background and enhancing the structure details. Although the above methods employ attention to combine multi-scale features, thereby avoiding equal weighted representation of important features at different scales, the independently generated attention maps at different scales are still insufficient in the mutual response of multi-scale features.

Therefore, we design the first module of the global pyramid attention unit, the pair-wise pyramid attention module, to perform adjacent attention interaction in a multi-scale pyramidal structure between adjacent scales, rather than learning all attention masks in a fixed set of scales. In other words, the attention between features of adjacent scales is correlated in pairs to complete the information interaction between different scales, and gradually refine and restore the top-level features to weaken the influence of high-similarity confusing categories. Specifically, as shown in Fig. \ref{3}, in order to obtain multi-scale pyramidal features, we upsample the top-level feature ${f_{top}}$ to 2x, 4x and 8x by bilinear interpolation, thereby generating the multi-scale representation features ${f_2}$, ${f_4}$ and ${f_8}$. We construct such a pyramidal feature ${f_r} \in {R^{(C/r) \times \,(rW) \times \,(rH)}}$, where $r \in {\rm{\{ }}2,4,8{\rm{\} }}$ indexes the pyramid scales. Then, pair-wise combinations are performed according to the scales from large to small, i.e., starting from ${f_4}$ and ${f_8}$, the higher scales are prioritized because they have more global contextual information. Subsequently, we extract the attention mask from the lower scale feature map ${f_4}$ in the adjacent scales, and then the mined attention mask is applied to the corresponding higher scale feature map ${f_8}$, so that we can choose which predictions need to be improved by higher scales. Relatively, we extract the attention mask from the neighboring scale high scale feature map ${f_8}$ similarly and then apply the phase multiplication over the low scale feature map ${f_4}$. Therefore, the two relative attentional streams complement mutually and highlight the different scale features'interaction, making the different scale features more explicitly accompanied with each other's important information and fused. Finally, we concatenate the outputs to generate a first cascade feature ${F_{1}}$. By parity of reasoning, cycling the above description to obtain the intermediate scale cascade result ${F_{2}}$ and the final cascade output ${F_{3}}$ step by step. The specific process can be expressed as follows:
\begin{equation}
	\alpha _{d}=Attn\left ( f_{d}\right )=\delta \circ conv\circ \sigma \circ upsample\left ( f_{d}\right )
\end{equation}
\begin{equation}
	\alpha _{2d}=Attn\left ( f_{2d}\right )=\delta \circ conv\circ \sigma \circ deconv\left ( f_{2d}\right )
\end{equation}
\begin{equation}
	F_{t}=conv\left ( concat\left ( conv\left ( \left ( \alpha _{2d}\otimes f_{d}\right ),\left ( \alpha _{d}\otimes f_{2d} \right )\downarrow\right )\right )\right )
\end{equation}
where $t \in {\rm{\{ }}1,2,3{\rm{\} }}$, $d \in {\rm{\{ }}1,2,4{\rm{\} }}$, $conv$ represents a
convolution operation with the filter size of $1\times 1$, $concat$ represents feature channel concatenation, $\downarrow$ denotes 2x spatial down-sampling operation, $deconv$ represents deconvolution, $\sigma $ denotes the ReLU activation function, $\delta $ is the Sigmoid operation, and $ \circ $ denotes function composition.

\subsubsection{Global Context Modeling Module}
\ 
\newline
\indent Before feeding the refined result to the decoding part, we attempt to maintain the complete response of the entire structure of the hippocampal target. Inspired by [39], we introduce a second module of the global pyramid attention unit, the global context modeling module. It achieves feature adjustment and mutual reinforcement by connecting the global semantic relationships between pixel pairs for the feature map output $f \in {R^{C\, \times \,W\, \times \,H}}$ obtained from the last module, thus maintaining the integrity of the segmented target and yielding a better image rendering in the subsequent decoder. First, we are to evaluate the interactions between any two positions of the embedding vector. Then, by aggregating multiple local features to produce a global context-aware representation, such interactions can be formulated as a spatial correlation map ${f_c} \in {R^{P\, \times \,P}}$. Mathematically,
\begin{equation}{f_c} = {\{ {(R(\tilde f))^T} \odot R(\tilde f)\} ^T}\end{equation}
where ${\tilde f}$ represents top-level feature after normalizing, $R{\rm{(}} \cdot {\rm{)}}$ denotes dimension transformation operation that reshapes a matrix of $R^{D_{1}\times D_{2}\times D_{3}}$ into $R^{D_{1}\times D_{23}}$, ${D_{23}} = {D_2} \times {D_3}$ , $\odot$ denotes the matrix multiplication, and $P = H \times W$ counts the number of pixels. Thereafter, we can obtain an updated feature map ${f_g}$ that encodes global context relationships:
\begin{equation}{f_g} = {R^{ - 1}}(R(\tilde f) \odot {f_c})\end{equation}
where ${R^{ - 1}}$ is the inverse operator of $R{\rm{(}} \cdot {\rm{)}}$ that reshapes a matrix of $R^{D_{1}\times D_{23}}$ into $R^{D_{1}\times D_{2}\times D_{3}}$. Then, we integrate the updated feature into the original high-level feature $f$ in a residual connection manner to accomplish feature enhancement:
\begin{equation}
	{f^ * } = f \oplus \alpha \cdot (f \otimes {f_g}) 
\end{equation}
where $\alpha$ is a learnable weight parameter that controls the contribution of global contextual information, the integrated feature map ${f^ * }$ embeds global contextual dependencies, which can constrain the response consistency of the global regions.

\section{Experiments and Results}
In this section, we first introduce the benchmark datasets, and describe the implementation details. Then, we conduct experiments on these datasets and compare them with other excellent methods to evaluate the effectiveness of the proposed method.
\subsection{Datasets}
\subsubsection{Decathlon}
In the MICCAI decathlon challenge dataset for hippocampus head and body segmentation, it contains T1-weighted MPRAGE sequence structure MRI data from 263 subjects, and is manually labeled with left, right, and anterior-posterior hippocampal regions by Vanderbilt University Medical Center. We perform a random split of the entire dataset and use a ten-fold cross-validation approach, setting 10\% of the data as a test set for general performance evaluation, while all the remaining data are used to train and validate the network. Due to the huge differences in the number of hippocampal and background voxels in the MRIs, this leads to severe category imbalance. Therefore, we need to perform data preprocessing. Specifically, we resample all MRIs, align them to a common template space, and get a hippocampal region of interest (ROI) cube with a scale of 32$\times$32$\times$32. Among them, the common template space refers to each voxel that is aligned to the same size, i.e., we complete the voxel intensity normalization. Then, we selected 12 to 20 2D adjacent slices containing hippocampal information along the depth from the ROI cubes obtained from 263 subjects, which ensures the structural similarity of the slices and reduces the complexity of the computation. Finally, we perform data augmentation using random rotation, panning, random blurring, and channel thrashing to prevent overfitting.

\subsubsection{ADNI}
 The MRIs for hippocampal region segmentation in this experiment were obtained from the Alzheimer's Disease Neuroimaging Initiative (ADNI) database, which includes MR images of 135 patients and carries labels after manual segmentation. We used the images of 100 patients as the training set and the other 35 as the test set. The labeling of the hippocampal MR images obtained by manual segmentation was done by using the Hippocampal Harmonization Protocol (HHP), which is the accepted standard for hippocampal segmentation. All images contain a total of three cross-sections, axial, coronal and sagittal plane respectively. The ADNI data were divided into four groups according to clinical status, including AD patients, mild cognitive impairment converters MCI non-converters and healthy controls. Each voxel data size is 256$\times$256$\times$128. To reduce complexity and computational cost, in line with the processing of the Decathlon dataset, we normalize the voxel data of 135 patients to 128$\times$128$\times$128 and slice them for input into our two-dimensional neural network for learning, where the size of the input images are 128$\times$128$\times$1. Among the slices obtained, we select as many slices as possible containing the hippocampal region for better model convergence (189 slices for each patient, 378 slices for the left and right hippocampal images), including about 20 slices that do not show any tissue structure, which together formed our training and testing data. We first concatenate the left and right hippocampal images, and then performed weak data argumentation methods such as random flipping, panning and blurring on the slices for training.

\subsubsection{IBSR}
We also use the IBSR dataset provided by the Morphometric Analysis Center at Massachusetts General Hospital. Among others, it contains T1-weighted MR image data from different subjects with different resolutions (from 0.8$\times$0.8$\times$1.5 to 1.0$\times$1.0$\times$1.5mm), and the manual segmentation of the corresponding hippocampal region provided by the IBSR repository is used as the image label, where the hippocampus is considered as a homogeneous gray matter structure. These images have been spatially normalized to Talairach orientation (rotation only), The rest of the data processing operations are kept consistent with those of the Decathlon dataset. The age of the subjects range from young people less than 7 years old to 71 years old. Among these subjects, two-thirds are male and the remaining one-third are female. The volumes’ size is 256$\times$256$\times$128 voxels. Similarly, based on the total number of 2500 original images, data augmentation was performed by random flipping, panning, blurring and channel disruption to prevent overfitting.

\begin{table}[!t]
	\centering
	\scriptsize
	\caption{Performance comparison with different methods on the Decathlon dataset. The best result is in \textbf{bold}, the second best result is \underline{underlined}.}
	\begin{tabular}{p{60pt}p{25pt}p{30pt}p{30pt}p{40pt}}
		\toprule
		\makecell{\textbf{Methods}}& 
		\makecell{\textbf{DSC}($\uparrow$)}&
		\makecell{\textbf{mIoU}($\uparrow$)}& 
		\makecell{\textbf{Recall}($\uparrow$)}&
		\makecell{\textbf{Accuracy}($\uparrow$)}\\
		\midrule
		\makecell{Thyreau's method} & \makecell{.8155}& \makecell{.7860}& \makecell{.7982}& \makecell{.8325}\\
		\makecell{U-Net} & \makecell{.8425} & \makecell{.8170}& \makecell{.8304}& \makecell{.8600}	\\
		\makecell{U-Net++} & \makecell{.8660}& \makecell{.8453}& \makecell{.8566}& \makecell{.8820}\\
		\makecell{Folle's method}& \makecell{.8820}& \makecell{-}& \makecell{-}& \makecell{-} \\
		\makecell{Ataloglou's method}& \makecell{.8835} &\makecell{-}&\makecell{-}&\makecell{-}\\
		\makecell{Attention U-Net}& \makecell{.9064}& \makecell{.8826}& \makecell{\textbf{.9013}}& \makecell{\underline{.9367}} \\
		\makecell{nnU-Net}& \makecell{.9073}& \makecell{.8814}& \makecell{.8867}& \makecell{.9310}\\
		\makecell{Shi's method}& \makecell{\underline{.9245}}& \makecell{\underline{.8974}}& \makecell{.8825}& \makecell{.9341}\\
		\makecell{Our method}& \makecell{\textbf{.9311}}& \makecell{\textbf{.9001}}& \makecell{\textit{.8950}}& \makecell{\textbf{.9382}} \\
		\bottomrule
	\end{tabular}
	\label{t1}
\end{table}

\begin{table}[!t]
	\centering
	\scriptsize
	\caption{Performance comparison with different methods on the ADNI dataset. The best result is in \textbf{bold}, the second best result is \underline{underlined}.}
	\begin{tabular}{p{60pt}p{25pt}p{30pt}p{30pt}p{40pt}}
		\toprule
		\makecell{\textbf{Methods}}& 
		\makecell{\textbf{DSC}($\uparrow$)}&
		\makecell{\textbf{mIoU}($\uparrow$)}& 
		\textbf{Recall}($\uparrow$)&
		\textbf{Accuracy}($\uparrow$)\\
		\midrule
		\makecell{U-Net} & \makecell{.8344}& \makecell{.8395}& \makecell{.8600}& \makecell{.8631}\\
		\makecell{Thyreau's method} & \makecell{.8427}& \makecell{.8481}& \makecell{.8562}& \makecell{.8596}\\
		\makecell{Cao's method} & \makecell{.8563} & \makecell{-}& \makecell{-}& \makecell{-} \\
		\makecell{U-Net++} & \makecell{.8659}& \makecell{.8702}& \makecell{.8805}& \makecell{.8816}\\
		\makecell{Liu's method}& \makecell{.8700}& \makecell{-}& \makecell{-}& \makecell{-} \\
		\makecell{Attention U-Net}& \makecell{.9097}& \makecell{.9121}& \makecell{.9196}& \makecell{.9288} \\
		\makecell{nnU-Net}& \makecell{.9129}& \makecell{\underline{.9254}}& \makecell{.9237}& \makecell{.9268}\\
		\makecell{Shi's method}& \makecell{\underline{.9188}}& \makecell{.9249}& \makecell{\underline{.9259}}& \makecell{\textbf{.9373}}\\
		\makecell{{Our method}}& \makecell{\textbf{.9236}}& \makecell{\textbf{.9285}}& \makecell{\textbf{.9261}}& \makecell{\underline{.9350}} \\
		\bottomrule
	\end{tabular}
	\label{t2}
\end{table}

\begin{table}[!t]
	\centering
	\scriptsize
	\caption{Performance comparison with different methods on the IBSR dataset. The best result is in \textbf{bold}, the second best result is \underline{underlined}.}
	\begin{tabular}{p{60pt}p{25pt}p{30pt}p{30pt}p{40pt}}
		\toprule
		\makecell{\textbf{Methods}}& 
		\makecell{\textbf{DSC}($\uparrow$)}&
		\makecell{\textbf{mIoU}($\uparrow$)}& 
		\textbf{Recall}($\uparrow$)&
		\textbf{Accuracy}($\uparrow$)\\
		\midrule
		\makecell{Rousseau's method} & \makecell{.8300}& \makecell{-}& \makecell{-}& \makecell{-}\\
		\makecell{Zarpalas's method} & \makecell{.8520} & \makecell{-}& \makecell{-}& \makecell{-}	\\
		\makecell{U-Net} & \makecell{.8647}& \makecell{.8236}& \makecell{.8511}& \makecell{.8875}\\
		\makecell{Thyreau's method}& \makecell{.8680}& \makecell{.8400}& \makecell{.8437}& \makecell{.8755} \\
		\makecell{U-Net++}& \makecell{.8854}& \makecell{.8602}& \makecell{.8755}& \makecell{.9010} \\
		\makecell{David's method}& \makecell{.8972} &\makecell{-}&\makecell{-}&\makecell{-}\\
		\makecell{Attention U-Net}& \makecell{.9081}& \makecell{.8895}& \makecell{.8940}& \makecell{.9305} \\
		\makecell{nnU-Net}& \makecell{.9133}& \makecell{\textbf{.9124}}& \makecell{\underline{.8937}}& \makecell{.9330}\\
		\makecell{Shi's method}& \makecell{\underline{.9152}}& \makecell{.8960}& \makecell{.8923}& \makecell{\textbf{.9384}}\\
		\makecell{{Our method}}& \makecell{\textbf{.9193}}& \makecell{\underline{.8987}}& \makecell{\textbf{.9011}}& \makecell{\underline{.9380}} \\
		\bottomrule
	\end{tabular}
	\label{t3}
\end{table}

\subsection{Implementation Details and Evaluation method}
The proposed method is implemented on the Tensorflow backend using python 3.7 and the Keras library. We also implement our network by using the MindSpore Lite tool\footnote{\url{https://www.mindspore.cn/}}. During training, the initial weights are kept consistent throughout the network. We use Adam optimizer with a learning rate of $2.0\times{10^{ - 5}}$, binary cross-entropy as the loss function, and the batch size of the images is set to 32. After 300 epochs of iterations, the network gradually stabilizes. All the experiments are conducted in the environment of Windows 10-x64/GPU of NVIDIA Tesla K80.

For the acquisition and processing of hippocampal MRI data, we first extract slices of different cross-sections for each group of 3D scan and visualize all the slices. Through observation, we learn that in the stereoscopic structure of each 3D scan, the slice located in the middle part of the volume (about 20 slices) contains the most complete and accurate hippocampal volume (this is also in accordance with the imaging principle of MRI, the more at the edge of the data stereoscopic volume, the less brain structures are mapped), so we select the middle slices as our training and testing data. Secondly, for the slices that do not contain hippocampus volume or very little volume, i.e., slices with no visible targets, we randomly select one-third of the total data volume as noise data, which makes the model more robust in learning and training and effectively avoids over-fitting.
\begin{figure}[!t]
	\centering
	\includegraphics[width=\linewidth]{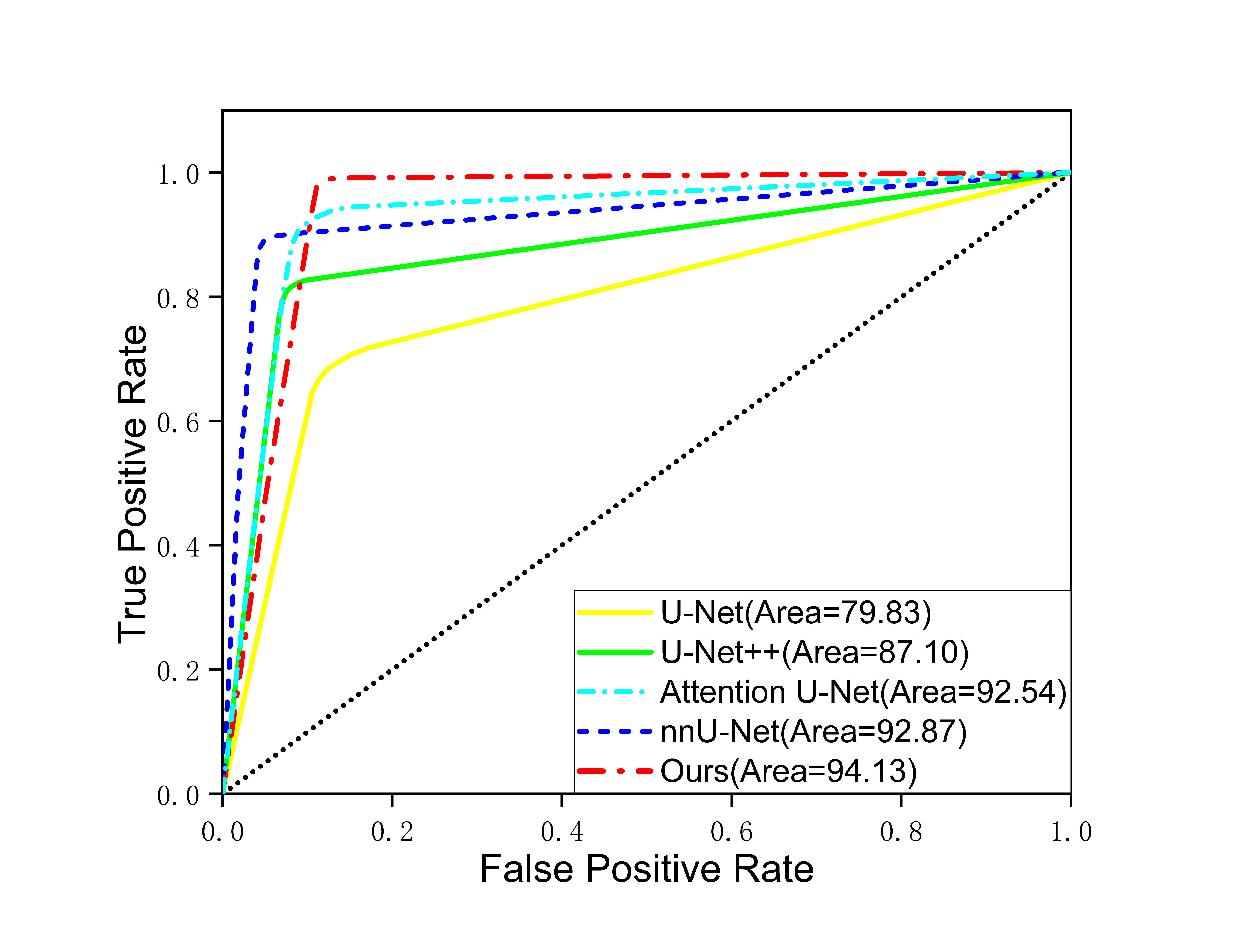}
	\caption{ROC curve of proposed models on the Decathlon.}
	\label{5}
\end{figure}
To evaluate our hippocampal segmentation method, we calculate four metrics including DSC, mIoU, Recall and Precision as evaluation metrics, and we use the class-balanced binary cross-entropy as the loss function for segmentation prediction supervision. Dice coefficient (DSC) and Intersection over Union (IoU) are the most commonly used metrics in image segmentation. We used DSC and mIou to compare the similarity between the generated hippocampal segmentation results and the original ground truth. We first embed the hippocampal slices into the model as a tensor, forward propagated the data in each epoch of each batch, calculate the Dice metric each time, and then back propagated to update the parameters. The mean Intersection over Union (mIoU) calculates the IoU for each semantic category of the image and computes the average of all categories. There is a correlation between DSC and mIoU, and we calculate that a more comprehensive analysis of the results can be provided by these two metrics for a better understanding of the results. In addition, the recall (true positive rate) metric, often used in the binary classification, can be used to calculate the ratio of correctly segmented pixels to the total number of pixels present in the ground truth. Precision (positive predictive value), on the other hand, can be interpreted as the ratio of the number of correctly segmented pixels to the total number of all pixels. They provide a good measure of the underfitting and overfitting present in the hippocampal segmentation. Finally, in addition to these four metrics, the receiver operating characteristic (ROC) curve analysis is also an important expression of performance in the binary classification.

\begin{figure*}
	\centering
	\includegraphics[width=\linewidth]{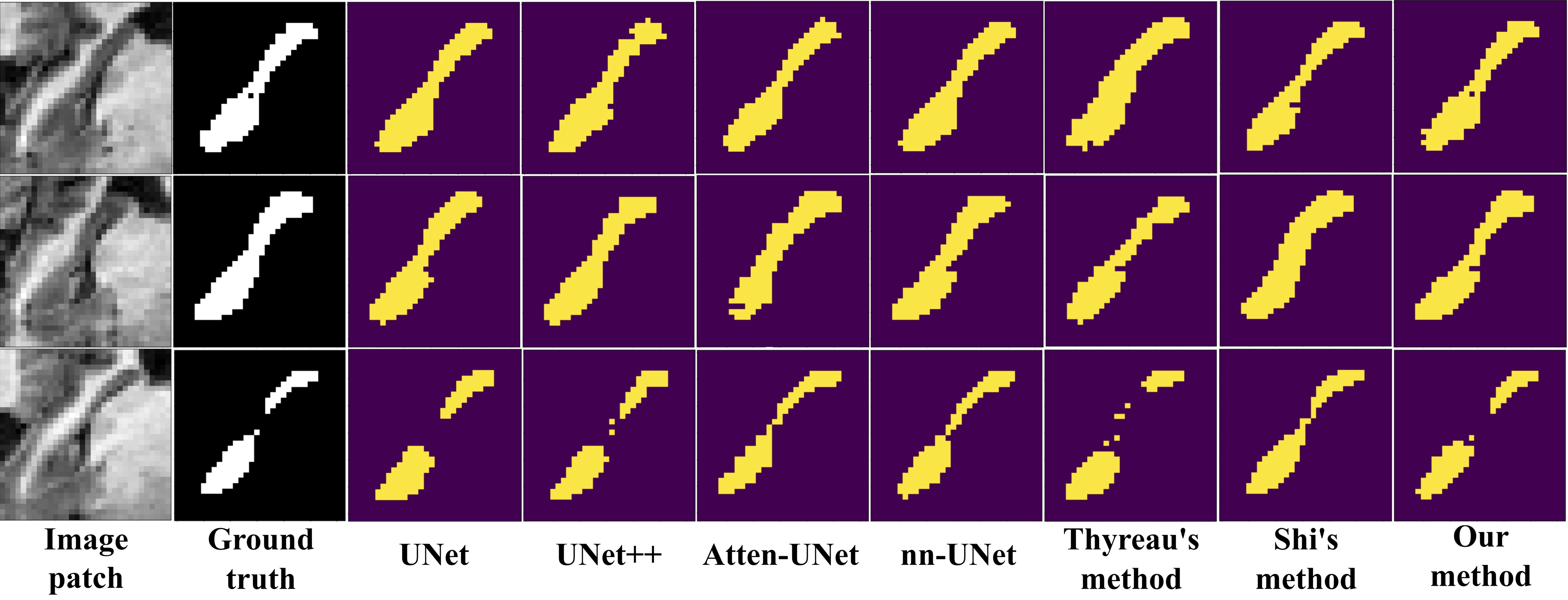}
	\caption{Visual comparisons of segmented hippocampal regions by different methods from the test data. In the case of various structural forms, it can be seen that our method generates segmentation maps do better in terms of key details and completeness of the hippocampus, and more closely approximate the real label.}
	\label{4}
\end{figure*}

\subsection{Compared with different methods}
\subsubsection{Quantitative Evaluation}
\ 
\newline
\indent We first compare the proposed model with other excellent semantic segmentation models on each publicly available dataset. They are mainly U-Net and its variants, which are representative in the field of medical image segmentation, including U-Net++ \cite{zhou2018unet++}, Attention U-Net \cite{oktay2018attention} and nnU-Net \cite{isensee2018nnu}. At the same time, we compare with other contributing hippocampus segmentation methods. Since multi-atlas methods are used in the field of medical image segmentation until now, the compared methods include contributing methods based on multi-atlas to deep neural networks. For the Decathlon dataset, the methods compared include Thyreau's method \cite{thyreau2018segmentation}, Ataloglou's method \cite{ataloglou2019fast}, Folle's method \cite{folle2019dilated} and Shi's method \cite{Shi2021DiscriminativeFN}. For the ADNI dataset, we additionally add Cao et al.'s method \cite{Cao2017MultitaskNN}, Liu et al.'s method \cite{Liu2020AMD} and Shi et al.'s method \cite{Shi2021DiscriminativeFN} as comparisons to enhance the validation of our proposed algorithm. For the IBSR dataset, the methods compared add Rousseau's method \cite{rousseau2011supervised}, Zarpalas's method \cite{zarpalas2014accurate}, David's method \cite{cardenas2019adaptive} and Shi's method \cite{Shi2021DiscriminativeFN}. Among them, the implementation codes of the methods are not open source except Thyreau et al. \cite{thyreau2018segmentation} and Shi et al. \cite{Shi2021DiscriminativeFN}, so the performance scores are taken from the corresponding papers. As shown in Tables \ref{t1}, \ref{t2} and \ref{t3}, they report the results of the quantitative evaluations of the four metric scores of DSC, mIoU, Recall and Accuracy on three different datasets, respectively. Obviously, from Tabel 1, on the Decathlon dataset, our proposed method achieves the best scores regarding DSC, mIoU and Accuracy with 93.11\%, 90.01\% and 93.82\%, respectively. Our proposed method with a percentage gain of 0.66\%, 0.27\% and 0.41\% over the second highest method of Shi et al. \cite{Shi2021DiscriminativeFN} in the three metrics of DSC, mIoU and Accuracy. Our method is only slightly inferior to Attention-UNet in the Recall metric at 89.50\%, but it also has the second highest performance score for that corresponding one. For the quantitative comparison on the ADNI reported in Table \ref{t2}, it can be found that, our method gained 0.48\%, 0.02\% percentage points in DSC and Recall metrics, respectively, compared to the second highest Shi et al. \cite{Shi2021DiscriminativeFN}, and 0.31\% percentage points in mIoU score compared to the second highest nn-UNet \cite{isensee2018nnu}.  The specific performance scores were 92.36\% on DSC, 92.85\% on mIoU, 92.61\% on Recall, and 93.50\% on Accuracy. Only in Accuracy score was 0.23\% percentage points lower than Shi et al. \cite{Shi2021DiscriminativeFN} in second place. From Table \ref{t3}, on the IBSR, we can see from that our proposed method is 0.41\% higher in the DSC metric than the second highest Shi et al. \cite{Shi2021DiscriminativeFN}, and 0.74\% higher in the Recall metric than the second highest nn-UNet \cite{isensee2018nnu}. Only the mIoU and Accuracy metrics succumbed to the second highest.

Finally, we present the ROC curves of all compared methods on the Decathlon dataset as shown in Fig. \ref{5}. It can be seen that the proposed method has a higher position on the ROC curve than the other methods and the value of 0.94 of the AUC corresponding to the curve is the maximum. In conclusion, our segmentation method ranks highest in the overall metrics. We experimentally demonstrated the effectiveness and utility of the model, as well as its usefulness and potential in the field of hippocampal segmentation domain.
\begin{figure*}[h]
	\centering
	\includegraphics[width=\linewidth]{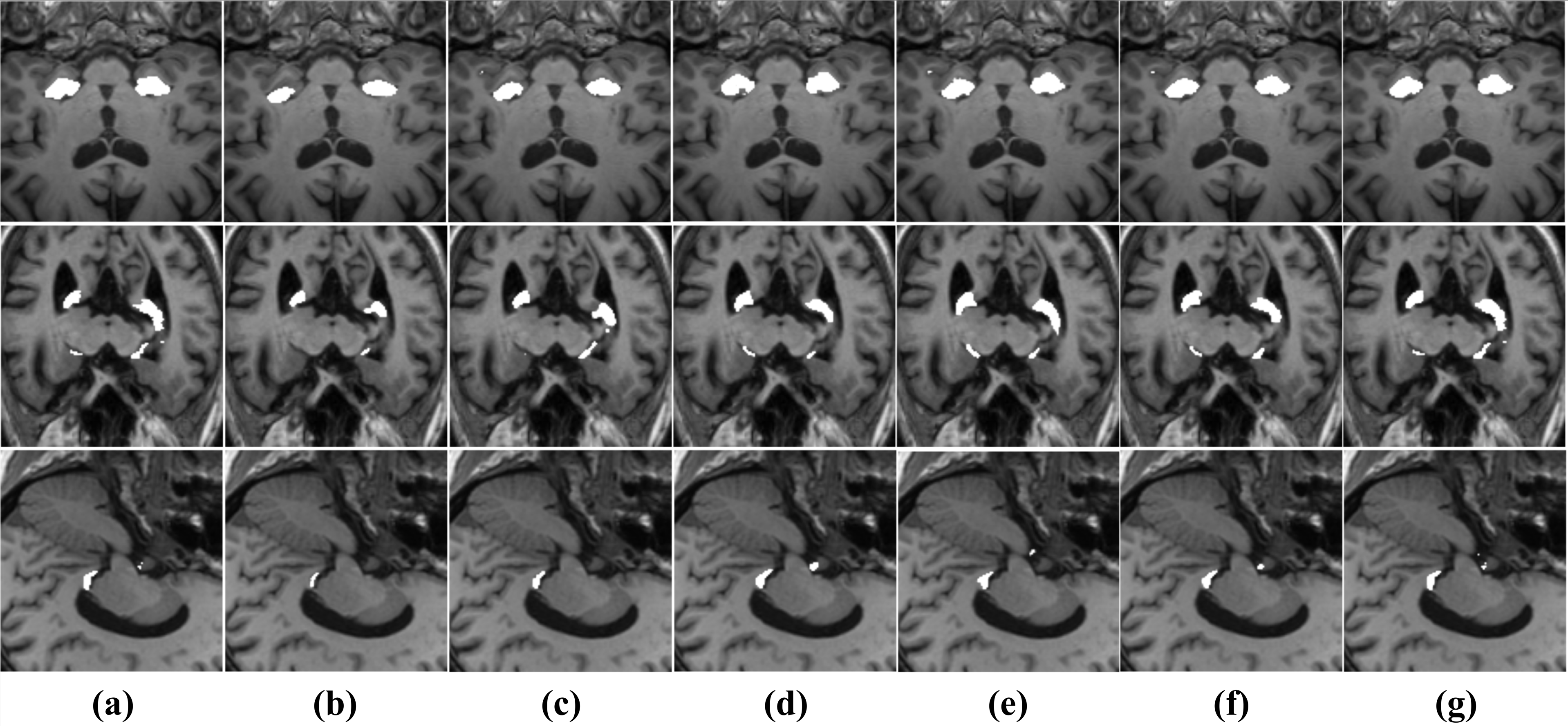}
	\caption{Visual comparisons of segmented hippocampal regions by different methods from the test data: (a) patches with segmentation labels, (b) visualization segmented patches of U-Net, (c) visualization segmented results of U-Net++, (d) visualization segmented patches of Attention-UNet, (e) visualization segmented patches of nn-UNet, (f) visualization segmented patches of Shi et al.'s method, (g) visualization segmented patches of our method. Shown from top row to bottom row are coronal, axial and sagittal plane respectively. As can be seen from the different perspectives of hippocampal patches, our proposed method is closer to the real label in terms of details and completeness of the target structure.}
	\label{6}
\end{figure*}
\subsubsection{Qualitative Evaluation}
\ 
\newline
\indent In addition, to further illustrate the advantages of the proposed method, we provide some visual examples of our model and all the comparison models, i.e., the semantic segmentation results of hippocampal MRIs. As shown in Fig. \ref{4}, from left to right, the original image, ground truth and the results of different models are shown. In terms of results, our proposed method also obtains significant improvements, making the segmentation more realistic and robust, and better in details. Note that our results are more robust to background/foreground interference (Fig. \ref{4}, third row), and capture more complete structural relationships in the hippocampal region (Fig. \ref{4}, first row), as well as present more details at the hippocampal head, trunk, and tail edges (Fig. \ref{4}, second row). This illustrates the mastery of the FAFC for discriminating features, the power of the FHA module and the global pyramid attention unit for the introduction of structural details and global contextual information. As can be seen from Fig. \ref{6}, the visualization results derived from hippocampal patches of different cross-sections, our method maintains more clarity, more accurate and complete view of the hippocampal regions compared to other different methods, again providing strong evidence of the robustness and effectiveness of our method.

\subsection{Ablation Study}
To validate the effectiveness of our proposed FAFC, FHA module and GPA unit in hippocampal segmentation, we perform ablation experiments.

As shown in Table \ref{t4}, we adopt U-Net that only concatenates high-level features after upsampling and low-level features as the baseline model, then add each module or unit progressively. It is worth stating that while we consider the GPA unit to be composed of the PPA and GCM modules together, they are inseparable. Each plays a different role in the overall globalized unit. However, to further refine the discussion of our method, in the ablation experiments, we have split the proposed GPA unit into a PPA module and a GCM module and discussed them separately. At this point, it is further divided into two cases, defined as 'w/o Gm, w/ Pm' and 'w/o Pm, w/ Gm', where Gm represents the GCM module and Pm represents the PPA module. From Table \ref{t4}, there is a significant performance improvement with the addition of different modules. First, the performance is improved by 3.98\% after adding the FAFC that contains only residual connection (only multi-level aggregation is performed without feature feedback) to form a chained structure. When we replace the residual connection in the chain structure with dense connection, i.e., after adding FAFC containing dense connection, the performance goes up by 0.71\%. This demonstrates the effectiveness of FAFC interoperability between different layers. Then, after adding FHA module on top of this, the DSC score improves to 91.50\%. Next, we can easily see that in the case of a GPA unit with only a PPA module on fine-grained progressive aggregation and no GCM module on global spatial attention, the performance gain is 0.99\% higher than the previous one, while the performance loss is 0.61\% lower compared to the full GPA unit with both sub-modules; on the contrary, it is slightly lower than the full GPA unit by 1.33\% performance score, while gaining 0.27\% performance gain than without the GPA unit. In a more detailed way, we have once again verified that the structure of our proposed algorithm fits reasonably well with each other and is indispensable to play an important role together for the hippocampus segmentation task.
{\footnotesize \begin{table*}[!htbp]
	\centering
	\caption{Quantitative evaluation results of different modules used in our proposed model, where ‘R’ means using residual connection between each FAFC, ‘D’ means using dense connection between each FAFC; 'w/o Gm, w/ Pm' refers to using only the PPA module and not the GCM module in the GPA unit; 'w/o Pm, w/ Gm' refers to using only the GCM module and not the PPA module in the GPA unit.}
	\begin{tabular}{|p{35pt}|p{40pt}|p{40pt}|p{20pt}|p{60pt}|p{60pt}|p{30pt}|p{20pt}|}
		\hline
		\makecell{\textbf{Baseline}} & \makecell{\textbf{FAFC(R)}}  &
		\makecell{\textbf{FAFC(D)}}  &
		\makecell{\textbf{FHA}}  &
		\makecell{\textbf{GPA} \\
		w/o Pm, w/ Gm}  &
		\makecell{\textbf{GPA} \\
		w/o Gm, w/ Pm}  &
		\makecell{\textbf{GPA} \\
		w/ both}  &
		\makecell{\textbf{DSC}} \\
		\hline
		\makecell{$\surd$} & & & & & & & \makecell{.8425} \\
		\makecell{$\surd$} & \makecell{$\surd$} & & & & & & \makecell{.8823} \\
		\makecell{$\surd$} & & \makecell{$\surd$} & & & & & \makecell{.8894} \\
		\makecell{$\surd$} & & \makecell{$\surd$} & \makecell{$\surd$} & & & & \makecell{.9150} \\
		\makecell{\textbf{$\surd$}} & & \makecell{\textbf{$\surd$}} & \makecell{\textbf{$\surd$}} & \makecell{\textbf{$\surd$}} & & & \makecell{.9249} \\
		\makecell{\textbf{$\surd$}} & & \makecell{\textbf{$\surd$}} & \makecell{\textbf{$\surd$}} & & \makecell{\textbf{$\surd$}} & & \makecell{.9283} \\
		\makecell{\textbf{$\surd$}} & & \makecell{\textbf{$\surd$}} & \makecell{\textbf{$\surd$}} & & & \makecell{\textbf{$\surd$}} &\makecell{.9311}\\
		\hline
	\end{tabular}
	\label{t4}
\end{table*}}

\begin{figure}[!t]
	\includegraphics[width=\linewidth]{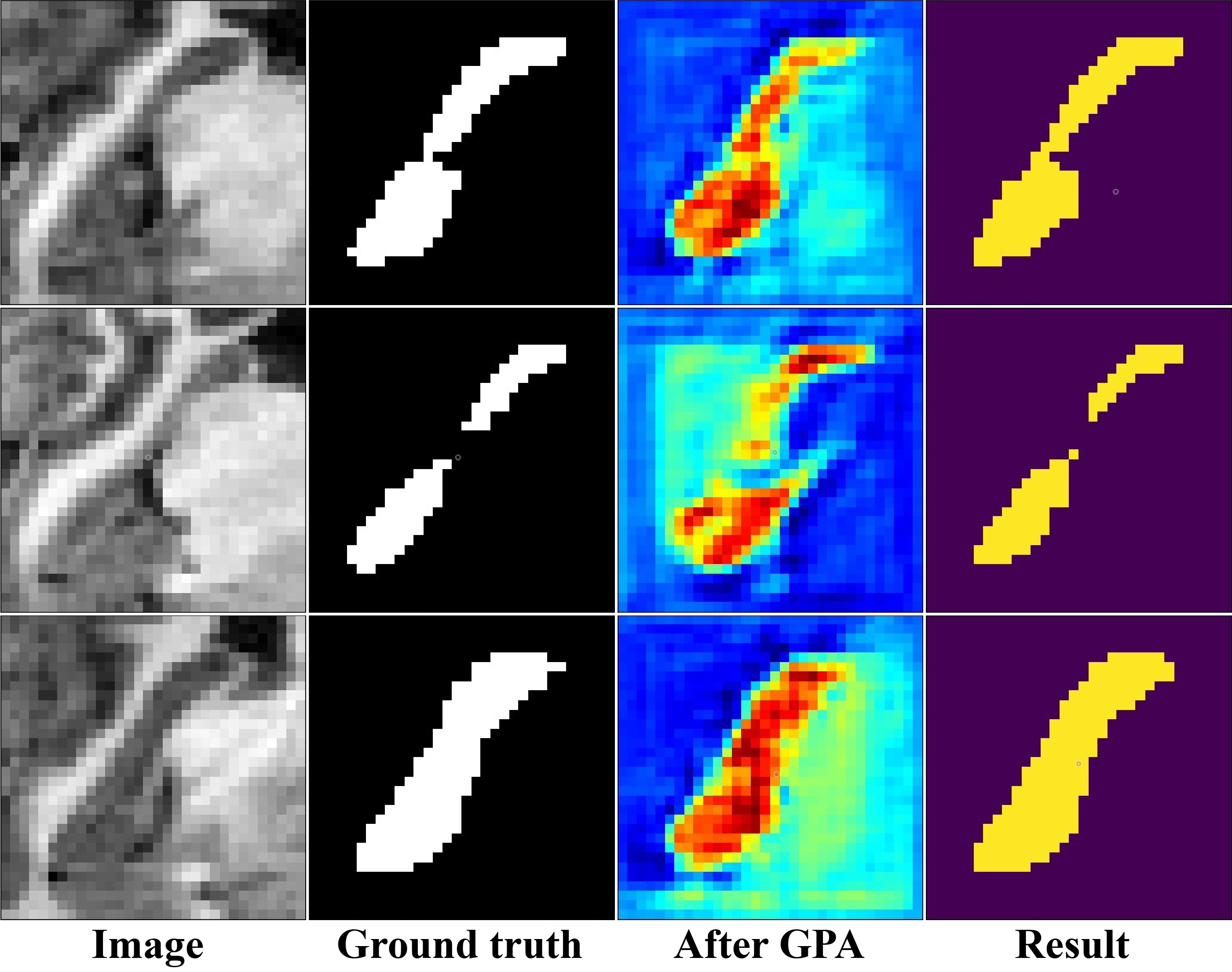}
	\caption{Visualization results after  global pyramid attention (GPA) unit. From left to right, we show an input image, the corresponding ground truth, the attention map, and the segmentation result.}
	\label{7}
\end{figure}

For the proposed global pyramid attention unit (GPA), at the middle of whole model, it up-bears the important features of rich semantic and spatial details, down-beats highlighting the global context and the fine-grained pixel recovery process. We believe that the unit is extremely critical in the whole learning and inference process. Therefore, we perform class activation mapping (CAM) on the feature map after the GPA unit to observe the effectiveness of the unit more specifically by way of attentional features. As shown in Fig. \ref{7}, the shades of color indicate the importance of each position to the classification. It can be seen that the region showing red color is clustered in the middle, overlapping with the region of interest in the hippocampus. In contrast, for the similarity category and background interference outside the segmentation target, it is reflected as the color around the red region gradually turns blue. The proposed global pyramid attention unit strongly divides the region of interest, the influence of the strong interference category and the background region into red, light blue and dark blue (Fig. \ref{7}, first row). Consistent with the ground truth, the global pyramid attention unit highlights the important long-range semantic information at this time, i.e., the regions of interest are clustered in the hippocampal region, in line with the answer. This fully verifies that the proposed unit has made a global response between the important regions and refined the high-level semantic information, thus achieving precise and complete segmentation. In summary, all the visual examples and quantitative results demonstrate the effectiveness of the key modules or units designed in our network.

\section{Conclusion}

In this paper, we propose a hierarchical feedback chain network for hippocampal segmentation. First, The feedback chain structure unit consists of the feature aggregation feedback chain and the feature handover attention module. The former learns deeper semantic features within the same level and aggregates wider cross-level features between different levels. In the latter, the selection and feedback of important features are implemented. Such chain units together with the backbone form the encoding part. Then, we bridge the global pyramid attention unit between the encoder to the decoder to further correct features before recovering resolution, including a pair-wise pyramid attention module that implements adjacent attention interactions and a global context modeling module that captures long-range relationships, thus improving the integrity of segmentation.


\bibliographystyle{IEEEtran}
\bibliography{sample-acmsmall-conf.bib}

\begin{thebibliography}{10}
\providecommand{\url}[1]{#1}
\csname url@samestyle\endcsname
\providecommand{\newblock}{\relax}
\providecommand{\bibinfo}[2]{#2}
\providecommand{\BIBentrySTDinterwordspacing}{\spaceskip=0pt\relax}
\providecommand{\BIBentryALTinterwordstretchfactor}{4}
\providecommand{\BIBentryALTinterwordspacing}{\spaceskip=\fontdimen2\font plus
\BIBentryALTinterwordstretchfactor\fontdimen3\font minus
  \fontdimen4\font\relax}
\providecommand{\BIBforeignlanguage}[2]{{%
\expandafter\ifx\csname l@#1\endcsname\relax
\typeout{** WARNING: IEEEtran.bst: No hyphenation pattern has been}%
\typeout{** loaded for the language `#1'. Using the pattern for}%
\typeout{** the default language instead.}%
\else
\language=\csname l@#1\endcsname
\fi
#2}}
\providecommand{\BIBdecl}{\relax}
\BIBdecl

\bibitem{prince2015world}
M.~J. Prince, A.~Wimo, M.~M. Guerchet, G.~C. Ali, Y.-T. Wu, and M.~Prina,
  ``World alzheimer report 2015-the global impact of dementia: An analysis of
  prevalence, incidence, cost and trends,'' 2015.

\bibitem{styner2004boundary}
M.~Styner, J.~A. Lieberman, D.~Pantazis, and G.~Gerig, ``Boundary and medial
  shape analysis of the hippocampus in schizophrenia,'' \emph{Med. Image
  Anal.}, vol.~8, no.~3, pp. 197--203, 2004.

\bibitem{tanveer2020machine}
M.~Tanveer, B.~Richhariya, R.~Khan, A.~Rashid, P.~Khanna, M.~Prasad, and
  C.~Lin, ``Machine learning techniques for the diagnosis of alzheimer’s
  disease: A review,'' \emph{ACM Trans. Multimedia Comput. Commun. Appl.},
  vol.~16, no.~1s, pp. 1--35, 2020.

\bibitem{bremner2000hippocampal}
J.~D. Bremner, M.~Narayan, E.~R. Anderson, L.~H. Staib, H.~L. Miller, and D.~S.
  Charney, ``Hippocampal volume reduction in major depression,'' \emph{American
  Journal of Psychiatry}, vol. 157, no.~1, pp. 115--118, 2000.

\bibitem{madan2017emotional}
C.~R. Madan, E.~Fujiwara, J.~B. Caplan, and T.~Sommer, ``Emotional arousal
  impairs association-memory: Roles of amygdala and hippocampus,''
  \emph{NeuroImage}, vol. 156, pp. 14--28, 2017.

\bibitem{saribudak2018gene}
A.~Saribudak, A.~A. Subick, N.~H. Kim, J.~A. Rutta, and M.~{\"U}. Uyar, ``Gene
  expressions, hippocampal volume loss, and mmse scores in computation of
  progression and pharmacologic therapy effects for alzheimer's disease,''
  \emph{IEEE/ACM Trans. Comput. Biol. Bioinform.}, vol.~17, no.~2, pp.
  608--622, 2018.

\bibitem{suk2015latent}
H.-I. Suk, S.-W. Lee, and D.~Shen, ``Latent feature representation with stacked
  auto-encoder for ad/mci diagnosis,'' \emph{Brain Structure and Function},
  vol. 220, no.~2, pp. 841--859, 2015.

\bibitem{wang2019hand}
B.~Wang, G.~Matcuk, and J.~Barbi{\v{c}}, ``Hand modeling and simulation using
  stabilized magnetic resonance imaging,'' \emph{ACM Trans. Graph.}, vol.~38,
  no.~4, pp. 1--14, 2019.

\bibitem{crm/tetci22/PSNet}
R.~Cong, W.~Song, J.~Lei, G.~Yue, Y.~Zhao, and S.~Kwong, ``{PSNet}: Parallel
  symmetric network for video salient object detection,'' \emph{IEEE Trans.
  Emerg. Topics Comput. Intell.}, early access, doi:
  10.1109/TETCI.2022.3220250.

\bibitem{litjens2017survey}
G.~Litjens, T.~Kooi, B.~E. Bejnordi, A.~A.~A. Setio, F.~Ciompi, M.~Ghafoorian,
  J.~A. Van Der~Laak, B.~Van~Ginneken, and C.~I. S{\'a}nchez, ``A survey on
  deep learning in med. image anal.'' \emph{Med. Image Anal.}, vol.~42, pp.
  60--88, 2017.

\bibitem{crm/acmmm21/bridgenet}
Q.~Tang, R.~Cong, R.~Sheng, L.~He, D.~Zhang, Y.~Zhao, and S.~Kwong,
  ``Bridgenet: {A} joint learning network of depth map super-resolution and
  monocular depth estimation,'' in \emph{Proc. ACM MM}, 2021, pp. 2148--2157.

\bibitem{crm/nips20/CoADNet}
Q.~Zhang, R.~Cong, J.~Hou, C.~Li, and Y.~Zhao, ``{CoADNet}: Collaborative
  aggregation-and-distribution networks for co-salient object detection,'' in
  \emph{Proc. NeurIPS}, 2020, pp. 6959--6970.

\bibitem{long2015fully}
J.~Long, E.~Shelhamer, and T.~Darrell, ``Fully convolutional networks for
  semantic segmentation,'' in \emph{Proc. CVPR}, 2015, pp. 3431--3440.

\bibitem{crm/tmm22/blindSR}
F.~Li, Y.~Wu, H.~Bai, W.~Lin, R.~Cong, and Y.~Zhao, ``Learning detail-structure
  alternative optimization for blind super-resolution,'' \emph{{IEEE} Trans.
  Multimedia}, early access, doi: 10.1109/TMM.2022.3152090.

\bibitem{zhou2020discriminative}
Y.~Zhou, L.~Liao, Y.~Gao, H.~Huang, and X.~Wei, ``A discriminative
  convolutional neural network with context-aware attention,'' \emph{ACM Trans.
  Intell. Syst. Technol.}, vol.~11, no.~5, pp. 1--21, 2020.

\bibitem{crm/aaai20/GCPANet}
Z.~Chen, Q.~Xu, R.~Cong, and Q.~Huang, ``Global context-aware progressive
  aggregation network for salient object detection,'' in \emph{Proc. AAAI},
  2020, pp. 10\,599--10\,606.

\bibitem{zotti2018convolutional}
C.~Zotti, Z.~Luo, A.~Lalande, and P.-M. Jodoin, ``Convolutional neural network
  with shape prior applied to cardiac mri segmentation,'' \emph{IEEE J. Biomed.
  Health Inform.}, vol.~23, no.~3, pp. 1119--1128, 2018.

\bibitem{crm/tip20/MCMT-GAN}
Y.~Huang, F.~Zheng, R.~Cong, W.~Huang, M.~R. Scott, and L.~Shao, ``{MCMT-GAN:}
  multi-task coherent modality transferable {GAN} for 3d brain image
  synthesis,'' \emph{{IEEE} Trans. Image Process.}, vol.~29, pp. 8187--8198,
  2020.

\bibitem{crm/acmmm20/NuI-Go}
C.~Li, H.~Fu, R.~Cong, Z.~Li, and Q.~Xu, ``{NuI-Go}: Recursive non-local
  encoder-decoder network for retinal image non-uniform illumination removal,''
  in \emph{Proc. ACM MM}, 2020, pp. 1478--1487.

\bibitem{lian2018multi}
C.~Lian, J.~Zhang, M.~Liu, X.~Zong, S.-C. Hung, W.~Lin, and D.~Shen,
  ``Multi-channel multi-scale fully convolutional network for 3d perivascular
  spaces segmentation in 7t mr images,'' \emph{Med. Image Anal.}, vol.~46, pp.
  106--117, 2018.

\bibitem{crm/tim22/covid}
R.~Cong, H.~Yang, Q.~Jiang, W.~Gao, H.~Li, C.~Wang, Y.~Zhao, and S.~Kwong,
  ``{BCS-Net}: Boundary, context, and semantic for automatic {COVID-19} lung
  infection segmentation from {CT} images,'' \emph{{IEEE} Trans. Instrum.
  Meas.}, vol.~71, pp. 1--11, 2022.

\bibitem{liu2020superpixel}
H.~Liu, H.~Wang, Y.~Wu, and L.~Xing, ``Superpixel region merging based on deep
  network for medical image segmentation,'' \emph{ACM Trans. Intell. Syst.
  Technol.}, vol.~11, no.~4, pp. 1--22, 2020.

\bibitem{crm/tce22/covid}
R.~Cong, Y.~Zhang, N.~Yang, H.~Li, X.~Zhang, R.~Li, Z.~Chen, Y.~Zhao, and
  S.~Kwong, ``Boundary guided semantic larning for real-time {COVID-19} lung
  infection segmentation system,'' \emph{IEEE Trans. Consum. Electron.},
  vol.~68, no.~4, pp. 376--386, 2022.

\bibitem{crm/jbhi22/polyp}
G.~Yue, W.~Han, B.~Jiang, T.~Zhou, R.~Cong, and T.~Wang, ``Boundary constraint
  network with cross layer feature integration for polyp segmentation,''
  \emph{IEEE J. Biomed. Health Inform.}, vol.~26, no.~8, pp. 4090--4099, 2022.

\bibitem{ronneberger2015u}
O.~Ronneberger, P.~Fischer, and T.~Brox, ``{U-Net}: Convolutional networks for
  biomedical image segmentation,'' in \emph{Proc. MICCAI}, 2015, pp. 234--241.

\bibitem{ibtehaz2020multiresunet}
N.~Ibtehaz and M.~S. Rahman, ``Multiresunet: Rethinking the u-net architecture
  for multimodal biomedical image segmentation,'' \emph{Neural Networks}, vol.
  121, pp. 74--87, 2020.

\bibitem{lin2021residual}
F.~Lin, W.~Zhou, J.~Deng, B.~Li, Y.~Lu, and H.~Li, ``Residual refinement
  network with attribute guidance for precise saliency detection,'' \emph{ACM
  Trans. Multimedia Comput. Commun. Appl.}, vol.~17, no.~3, pp. 1--19, 2021.

\bibitem{liu2018brain}
D.~Liu, H.~Zhang, M.~Zhao, X.~Yu, S.~Yao, and W.~Zhou, ``Brain tumor segmention
  based on dilated convolution refine networks,'' in \emph{IEEE 16th
  International Conference on Software Engineering Research, Management and
  Applications}.\hskip 1em plus 0.5em minus 0.4em\relax IEEE, 2018, pp.
  113--120.

\bibitem{punn2020inception}
N.~S. Punn and S.~Agarwal, ``Inception {U-Net} architecture for semantic
  segmentation to identify nuclei in microscopy cell images,'' \emph{ACM Trans.
  Multimedia Comput. Commun. Appl.}, vol.~16, no.~1, pp. 1--15, 2020.

\bibitem{li2018h}
X.~Li, H.~Chen, X.~Qi, Q.~Dou, C.-W. Fu, and P.-A. Heng, ``H-denseunet: hybrid
  densely connected unet for liver and tumor segmentation from ct volumes,''
  \emph{{IEEE} Trans. Med. Imag.}, vol.~37, no.~12, pp. 2663--2674, 2018.

\bibitem{yang2021densely}
Z.~Yang, P.~Xu, Y.~Yang, and B.-K. Bao, ``A densely connected network based on
  u-net for medical image segmentation,'' \emph{ACM Trans. Multimedia Comput.
  Commun. Appl.}, vol.~17, no.~3, pp. 1--14, 2021.

\bibitem{Zhu2019DilatedDU}
H.~Zhu, F.~Shi, L.~Wang, S.~C. Hung, M.-H. Chen, S.~Wang, W.~Lin, and D.~Shen,
  ``Dilated dense u-net for infant hippocampus subfield segmentation,''
  \emph{Frontiers in Neuroinformatics}, vol.~13, 2019.

\bibitem{Girum2021LearningWC}
K.~B. Girum, G.~Cr{\'e}hange, and A.~Lalande, ``Learning with context feedback
  loop for robust medical image segmentation,'' \emph{{IEEE} Trans. Med.
  Imag.}, vol.~40, pp. 1542--1554, 2021.

\bibitem{artaechevarria2009combination}
X.~Artaechevarria, A.~Munoz-Barrutia, and C.~Ortiz-de Solorzano, ``Combination
  strategies in multi-atlas image segmentation: application to brain mr data,''
  \emph{{IEEE} Trans. Med. Imag.}, vol.~28, no.~8, pp. 1266--1277, 2009.

\bibitem{rousseau2011supervised}
F.~Rousseau, P.~A. Habas, and C.~Studholme, ``A supervised patch-based approach
  for human brain labeling,'' \emph{{IEEE} Trans. Med. Imag.}, vol.~30, no.~10,
  pp. 1852--1862, 2011.

\bibitem{zarpalas2014accurate}
D.~Zarpalas, P.~Gkontra, P.~Daras, and N.~Maglaveras, ``Accurate and fully
  automatic hippocampus segmentation using subject-specific 3d optimal local
  maps into a hybrid active contour model,'' \emph{IEEE J. Transl. Eng. Health
  Med.}, vol.~2, pp. 1--16, 2014.

\bibitem{cardenas2019adaptive}
D.~C{\'a}rdenas-Pe{\~n}a, A.~Tobar-Rodr{\'\i}guez, G.~Castellanos-Dominguez,
  A.~D.~N. Initiative \emph{et~al.}, ``Adaptive bayesian label fusion using
  kernel-based similarity metrics in hippocampus segmentation,'' \emph{Journal
  of Medical Imaging}, vol.~6, no.~1, p. 014003, 2019.

\bibitem{fischl2012freesurfer}
B.~Fischl, ``Freesurfer,'' \emph{Neuroimage}, vol.~62, no.~2, pp. 774--781,
  2012.

\bibitem{thyreau2018segmentation}
B.~Thyreau, K.~Sato, H.~Fukuda, and Y.~Taki, ``Segmentation of the hippocampus
  by transferring algorithmic knowledge for large cohort processing,''
  \emph{Med. Image Anal.}, vol.~43, pp. 214--228, 2018.

\bibitem{ataloglou2019fast}
D.~Ataloglou, A.~Dimou, D.~Zarpalas, and P.~Daras, ``Fast and precise
  hippocampus segmentation through deep convolutional neural network ensembles
  and transfer learning,'' \emph{Neuroinformatics}, vol.~17, no.~4, pp.
  563--582, 2019.

\bibitem{folle2019dilated}
L.~Folle, S.~Vesal, N.~Ravikumar, and A.~Maier, ``Dilated deeply supervised
  networks for hippocampus segmentation in {MRI},'' in \emph{Bildverarbeitung
  f{\"u}r die Medizin}, 2019, pp. 68--73.

\bibitem{cui2018hippocampus}
R.~Cui and M.~Liu, ``Hippocampus analysis by combination of 3-d densenet and
  shapes for alzheimer's disease diagnosis,'' \emph{IEEE J. Biomed. Health
  Inform.}, vol.~23, no.~5, pp. 2099--2107, 2018.

\bibitem{Cao2017MultitaskNN}
L.-L. Cao, L.~Li, J.~Zheng, X.~Fan, F.~Yin, H.~Shen, and J.~Zhang, ``Multi-task
  neural networks for joint hippocampus segmentation and clinical score
  regression,'' \emph{Multimedia Tools and Applications}, vol.~77, pp.
  29\,669--29\,686, 2017.

\bibitem{Liu2020AMD}
M.~Liu, F.~Li, H.~Yan, K.~Wang, Y.~Ma, L.~Shen, and M.~Xu, ``A multi-model deep
  convolutional neural network for automatic hippocampus segmentation and
  classification in alzheimer’s disease,'' \emph{NeuroImage}, vol. 208, 2020.

\bibitem{Shi2021DiscriminativeFN}
J.~Shi, R.-R. Zhang, L.~Guo, L.~Gao, H.~Ma, and J.~Wang, ``Discriminative
  feature network based on a hierarchical attention mechanism for semantic
  hippocampus segmentation,'' \emph{IEEE J. Biomed. Health Inform.}, vol.~25,
  pp. 504--513, 2021.

\bibitem{he2019dynamic}
J.~He, Z.~Deng, and Y.~Qiao, ``Dynamic multi-scale filters for semantic
  segmentation,'' in \emph{Proc. ICCV}, 2019, pp. 3562--3572.

\bibitem{crm/tgrs22/RRNet}
R.~Cong, Y.~Zhang, L.~Fang, J.~Li, Y.~Zhao, and S.~Kwong, ``{RRNet}: Relational
  reasoning network with parallel multi-scale attention for salient object
  detection in optical remote sensing images,'' \emph{IEEE Trans. Geosci.
  Remote Sens.}, vol.~60, pp. 1558--0644, 2022.

\bibitem{he2019adaptive}
J.~He, Z.~Deng, L.~Zhou, Y.~Wang, and Y.~Qiao, ``Adaptive pyramid context
  network for semantic segmentation,'' in \emph{Proc. CVPR}, 2019, pp.
  7519--7528.

\bibitem{crm/tcyb22/rsi}
X.~Zhou, K.~Shen, L.~Weng, R.~Cong, B.~Zheng, J.~Zhang, and C.~Yan,
  ``Edge-guided recurrent positioning network for salient object detection in
  optical remote sensing images,'' \emph{IEEE Trans. Cybern.}, early access,
  doi: 10.1109/TCYB.2022.3163152.

\bibitem{kirillov2019panoptic}
A.~Kirillov, R.~Girshick, K.~He, and P.~Doll{\'a}r, ``Panoptic feature pyramid
  networks,'' in \emph{Proc. CVPR}, 2019, pp. 6399--6408.

\bibitem{crm/tip21/DAFNet}
Q.~Zhang \emph{et~al.}, ``Dense attention fluid network for salient object
  detection in optical remote sensing images,'' \emph{IEEE Trans. Image
  Process.}, vol.~30, pp. 1305--1317, 2021.

\bibitem{zhu2019asymmetric}
Z.~Zhu, M.~Xu, S.~Bai, T.~Huang, and X.~Bai, ``Asymmetric non-local neural
  networks for semantic segmentation,'' in \emph{Proc. ICCV}, 2019, pp.
  593--602.

\bibitem{yuan2019spatial}
Y.~Yuan, J.~Fang, X.~Lu, and Y.~Feng, ``Spatial structure preserving feature
  pyramid network for semantic image segmentation,'' \emph{ACM Trans.
  Multimedia Comput. Commun. Appl.}, vol.~15, no.~3, pp. 1--19, 2019.

\bibitem{crm/tgrs19/rsi}
C.~Li, R.~Cong, J.~Hou, S.~Zhang, Y.~Qian, and S.~Kwong, ``Nested network with
  two-stream pyramid for salient object detection in optical remote sensing
  images,'' \emph{IEEE Trans. Geosci. Remote Sens.}, vol.~57, no.~11, pp.
  9156--9166, 2019.

\bibitem{zhao2017pyramid}
H.~Zhao, J.~Shi, X.~Qi, X.~Wang, and J.~Jia, ``Pyramid scene parsing network,''
  in \emph{Proc. CVPR}, 2017, pp. 2881--2890.

\bibitem{chen2017deeplab}
L.-C. Chen, G.~Papandreou, I.~Kokkinos, K.~Murphy, and A.~L. Yuille, ``Deeplab:
  Semantic image segmentation with deep convolutional nets, atrous convolution,
  and fully connected crfs,'' \emph{IEEE Trans. Pattern Anal. Mach. Intell.},
  vol.~40, no.~4, pp. 834--848, 2017.

\bibitem{chen2017rethinking}
L.-C. Chen, G.~Papandreou, F.~Schroff, and H.~Adam, ``Rethinking atrous
  convolution for semantic image segmentation,'' \emph{arXiv preprint
  arXiv:1706.05587}, 2017.

\bibitem{zhao2018icnet}
H.~Zhao, X.~Qi, X.~Shen, J.~Shi, and J.~Jia, ``Icnet for real-time semantic
  segmentation on high-resolution images,'' in \emph{Proceedings of the
  European Conference on Computer Vision}, 2018, pp. 405--420.

\bibitem{crm/tip22/CIRNet}
R.~Cong, Q.~Lin, C.~Zhang, C.~Li, X.~Cao, Q.~Huang, and Y.~Zhao, ``{CIR-Net}:
  Cross-modality interaction and refinement for {RGB-D} salient object
  detection,'' \emph{IEEE Trans. Image Process.}, vol.~31, pp. 6800--6815,
  2022.

\bibitem{crm/tip21/DynamicRGBDSOD}
H.~Wen, C.~Yan, X.~Zhou, R.~Cong, Y.~Sun, B.~Zheng, J.~Zhang, Y.~Bao, and
  G.~Ding, ``Dynamic selective network for {RGB-D} salient object detection,''
  \emph{{IEEE} Trans. Image Process.}, vol.~30, pp. 9179--9192, 2021.

\bibitem{crm/tcyb22/glnet}
R.~Cong, N.~Yang, C.~Li, H.~Fu, Y.~Zhao, Q.~Huang, and S.~Kwong,
  ``Global-and-local collaborative learning for co-salient object detection,''
  \emph{IEEE Trans. Cybern.}, early access, doi: 10.1109/TCYB.2022.3169431.

\bibitem{crm/tmm22/TNet}
R.~Cong, K.~Zhang, C.~Zhang, F.~Zheng, Y.~Zhao, Q.~Huang, and S.~Kwong, ``Does
  thermal really always matter for {RGB-T} salient object detection?''
  \emph{IEEE Trans. Multimedia}, early access, doi: 10.1109/TMM.2022.3216476.

\bibitem{crm/tcsvt22/weaklySOD}
R.~Cong, Q.~Qin, C.~Zhang, Q.~Jiang, S.~Wang, Y.~Zhao, and S.~Kwong, ``A weakly
  supervised learning framework for salient object detection via hybrid
  labels,'' \emph{IEEE Trans. Circuits Syst. Video Technol.}, early access,
  doi: 10.1109/TCSVT.2022.3205182.

\bibitem{crm/acmmm21/CDINet}
C.~Zhang, R.~Cong, Q.~Lin, L.~Ma, F.~Li, Y.~Zhao, and S.~Kwong,
  ``Cross-modality discrepant interaction network for {RGB-D} salient object
  detection,'' in \emph{Proc. ACM MM}, 2021, pp. 2094--2102.

\bibitem{crm/CVPR21/depthSR}
L.~He, H.~Zhu, F.~Li, H.~Bai, R.~Cong, C.~Zhang, C.~Lin, M.~Liu, and Y.~Zhao,
  ``Towards fast and accurate real-world depth super-resolution: Benchmark
  dataset and baseline,'' in \emph{Proc. {IEEE} CVPR}, 2021, pp. 9229--9238.

\bibitem{hu2018squeeze}
J.~Hu, L.~Shen, and G.~Sun, ``Squeeze-and-excitation networks,'' in \emph{Proc.
  CVPR}, 2018, pp. 7132--7141.

\bibitem{fu2019dual}
J.~Fu, J.~Liu, H.~Tian, Y.~Li, Y.~Bao, Z.~Fang, and H.~Lu, ``Dual attention
  network for scene segmentation,'' in \emph{Proc. CVPR}, 2019, pp. 3146--3154.

\bibitem{woo2018cbam}
S.~Woo, J.~Park, J.-Y. Lee, and I.~S. Kweon, ``Cbam: Convolutional block
  attention module,'' in \emph{Proceedings of the European Conference on
  Computer Vision}, 2018, pp. 3--19.

\bibitem{huang2017densely}
G.~Huang, Z.~Liu, L.~Van Der~Maaten, and K.~Q. Weinberger, ``Densely connected
  convolutional networks,'' in \emph{Proc. CVPR}, 2017, pp. 4700--4708.

\bibitem{zhang2020dense}
Q.~Zhang, R.~Cong, C.~Li, M.-M. Cheng, Y.~Fang, X.~Cao, Y.~Zhao, and S.~Kwong,
  ``Dense attention fluid network for salient object detection in optical
  remote sensing images,'' \emph{IEEE Trans. Image Process.}, vol.~30, pp.
  1305--1317, 2020.

\bibitem{wang2019salient}
W.~Wang, S.~Zhao, J.~Shen, S.~C. Hoi, and A.~Borji, ``Salient object detection
  with pyramid attention and salient edges,'' in \emph{Proc. CVPR}, 2019, pp.
  1448--1457.

\bibitem{zhou2018unet++}
Z.~Zhou, M.~M.~R. Siddiquee, N.~Tajbakhsh, and J.~Liang, ``Unet++: A nested
  u-net architecture for medical image segmentation,'' in \emph{Deep learning
  in Med. Image Anal. and Multimodal Learning for Clinical Decision Support},
  2018, pp. 3--11.

\bibitem{oktay2018attention}
O.~Oktay, J.~Schlemper, L.~L. Folgoc, M.~Lee, M.~Heinrich, K.~Misawa, K.~Mori,
  S.~McDonagh, N.~Y. Hammerla, B.~Kainz \emph{et~al.}, ``Attention u-net:
  Learning where to look for the pancreas,'' \emph{arXiv preprint
  arXiv:1804.03999}, 2018.

\bibitem{isensee2018nnu}
F.~Isensee, J.~Petersen, A.~Klein, D.~Zimmerer, P.~F. Jaeger, S.~Kohl,
  J.~Wasserthal, G.~Koehler, T.~Norajitra, S.~Wirkert \emph{et~al.}, ``nnu-net:
  Self-adapting framework for u-net-based medical image segmentation,''
  \emph{arXiv preprint arXiv:1809.10486}, 2018.

\end{thebibliography}

\end{document}